# Brain Tumor Segmentation and Radiomics Survival Prediction: Contribution to the BRATS 2017 Challenge

Fabian Isensee[1], Philipp Kickingereder[2], Wolfgang Wick[3], Martin Bendszus[2], and Klaus H. Maier-Hein[1]

[1] Division of Medical Image Computing, German Cancer Research Center (DKFZ), Heidelberg, Germany
[2] Department of Neuroradiology, Heidelberg University Hospital, Heidelberg, Germany
[3] Neurology Clinic, Heidelberg University Hospital, Heidelberg, Germany

**Abstract.** Quantitative analysis of brain tumors is critical for clinical decision making. While manual segmentation is tedious, time consuming and subjective, this task is at the same time very challenging to solve for automatic segmentation methods. In this paper we present our most recent effort on developing a robust segmentation algorithm in the form of a convolutional neural network. Our network architecture was inspired by the popular U-Net and has been carefully modified to maximize brain tumor segmentation performance. We use a dice loss function to cope with class imbalances and use extensive data augmentation to successfully prevent overfitting. Our method beats the current state of the art on BraTS 2015, is one of the leading methods on the BraTS 2017 validation set (dice scores of 0.896, 0.797 and 0.732 for whole tumor, tumor core and enhancing tumor, respectively) and achieves very good Dice scores on the test set (0.858 for whole, 0.775 for core and 0.647 for enhancing tumor). We furthermore take part in the survival prediction subchallenge by training an ensemble of a random forest regressor and multilayer perceptrons on shape features describing the tumor subregions. Our approach achieves 52.6% accuracy, a Spearman correlation coefficient of 0.496 and a mean square error of 209607 on the test set.

## 1 Introduction

Quantitative assessment of brain tumors provides valuable information and therefore constitutes an essential part of diagnostic procedures. Automatic segmentation is attractive in this context, as it allows for faster, more objective and potentially more accurate description of relevant tumor parameters, such as the volume of its subregions. Due to the irregular nature of tumors, however, the development of algorithms capable of automatic segmentation remains challenging.

The brain tumor segmentation challenge (BraTS) [1] aims at encouraging the development of state of the art methods for tumor segmentation by providing a large dataset of annotated low grade gliomas (LGG) and high grade glioblastomas (HGG). Unlike the previous years, the BraTS 2017 training dataset, which consists of 210 HGG and 75 LGG cases, was annotated manually by one to four raters and all segmentations were approved by expert raters [2–4]. For each patient a T1 weighted, a post-contrast T1-weighted, a T2-weighted and a FLAIR MRI was provided. The MRI originate from 19 institutions and were acquired with different protocols, magnetic field strengths and MRI scanners. Each tumor was segmented into edema (label 2), necrosis and non-enhancing tumor (label 1) and active/enhancing tumor (label 4). The segmentation performance of participating algorithms is measured based on the DICE coefficient, sensitivity, specificity and Hausdorff distance. Additional to the segmentation challenge, BraTS 2017 also required participants to develop an algorithm for survival prediction. For this purpose the survival (in days) of 163 training cases was provided as well.

Inspired by the recent success of convolutional neural networks, an increasing number of deep learning based automatic segmentation algorithms have been proposed. Havaei et al. [5] use a multi-scale architecture by combining features from pathways with different filter sizes. They furthermore improve their results by cascading their models. Pereira et al. [6] stack more convolutional layers with smaller (3x3) filter sizes. They develop separate networks for segmenting low grade and high grade glioblastomas (LGG and HGG, respectively). Their LGG network consists of 4 convolutional layers, followed by two dense and a classification network. The HGG network is composed of 7 convolutional layers. Both [5] and [6] use 2D convolutions. Kamnitsas et al. [7] proposed a fully connected multi-scale CNN that was among the first to employ 3D convolutions. It comprises a high resolution and a low resolution pathway that are recombined to form the final segmentation output. For their submission to the brain tumor segmentation challenge in 2016 [8], they enhanced their architecture through the addition of residual connections, yielding minor improvements. They addressed the class imbalance problem through a sophisticated training data sampling strategy. Kayalibay et al. [9] developed very successful adaptation of the popular U-Net architecture [10] and achieved state of the art results for the BraTS 2015 dataset. Notably, they employed a Jaccard loss function that intrinsically handles class imbalances. They make use of the large receptive field of their architecture to process entire patients at once, at the cost of being able to train with only one patient per batch.

Here we propose our contribution to the BraTS 2017 challenge that is also based on the popular U-Net architecture [10]. Being both based on the U-Net, our network architecture shares some similarities with [9]. However, there are a multitude of different design choices that me made regarding the exact architecture of the context pathway, normalization schemes, number of feature maps throughout the network, nonlinearity and the structure of the upsampling pathway. Particularly through optimizing the number of feature maps in the lo-calization pathway, our network uses twice as many filters than [9] while being trained with only a slightly smaller input patch size and a larger batch size. We furthermore employ a multiclass adaptation of the dice loss [11] and make extensive use of data augmentation.

Image based tumor phenotyping and derived clinically relevant parameters such as predicted survival is typically done by means of radiomics. Intensity, shape and texture features are thereby computed from segmentation masks of the tumor subregions and subsequently used to train a machine learning algorithm. These features may also be complemented by other measures handcrafted to the problem at hand, such as the distance of the tumor to the ventricles and critical structures in the brain [12]. Although our main focus was put on the segmentation part of the challenge, we developed a simple radiomics based approach combined with a random forest regressor and a multilayer perceptron ensemble for survival prediction.

## 2 Methods

### 2.1 Segmentation

**Data preprocessing** With MRI intensity values being non standardized, normalization is critical to allow for data from different institutes, scanners and acquired with varying protocols to be processed by one single algorithm. This is particularly true for neural networks where imaging modalities are typically treated as color channels. Here we need to ensure that the value ranges match not only between patients but between the modalities as well in order to avoid initial biases of the network. We found the following simple workflow to work surprisingly well. First, we normalize each modality of each patient independently by subtracting the mean and dividing by the standard deviation of the brain region. We then clip the resulting images at $[-5, 5]$ to remove outliers and subsequently rescale to $[0, 1]$, with the non-brain region being set to 0.

**Network architecture** Our network is inspired by the U-Net architecture [10]. We designed the network to process large 3D input blocks of 128x128x128 voxels. In contrast to many previous approaches who manually combined different input resolutions or pathways with varying filter sizes, the U-Net based approach allows the network to intrinsically recombine different scales throughout the entire network.

Just like the U-Net, our architecture comprises a context aggregation pathway that encodes increasingly abstract representations of the input as we progress deeper into the network, followed by a localization pathway that recombines these representations with shallower features to precisely localize the structures of interest. We refer to the vertical depth (the depth in the U shape) as level, with higher levels being lower spatial resolution, but higher dimensional feature representations.

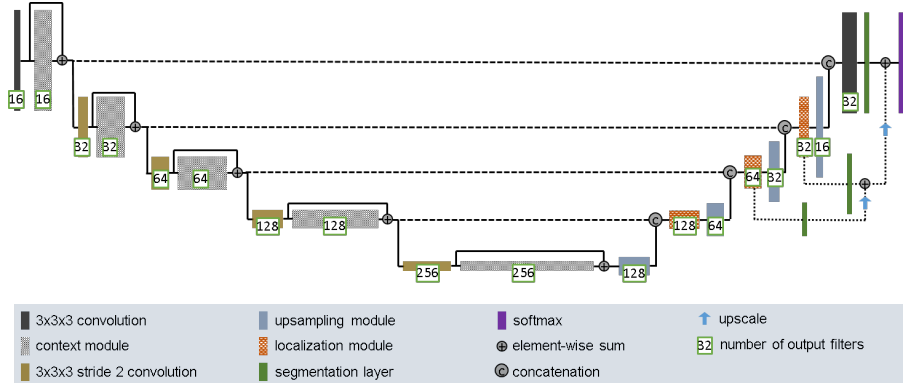

**Fig. 1.** Network architecture. Our architecture is derived from the U-Net [10]. The context pathway (left) aggregates high level information that is subsequently localized precisely in the localization pathway (right). Inspired by [9] we inject gradient signals deep into the network through deep supervision.

The activations in the context pathway are computed by context modules. Each context module is in fact a pre-activation residual block [13] with two 3x3x3 convolutional layers and a dropout layer ($p_{\text{drop}} = 0.3$) in between. Context modules are connected by 3x3x3 convolutions with input stride 2 to reduce the resolution of the feature maps and allow for more features while descending down the aggregation pathway.

As stated previously, the localization pathway is designed to take features from lower levels of the network that encode contextual information at low spatial resolution and transfer that information to a higher spatial resolution. This is achieved by first upsampling the low resolution feature maps, which is done by means of a simple upscale that repeats the feature voxels twice in each spatial dimension, followed by a 3x3x3 convolution that halves the number of feature maps. Compared to the more frequently employed transposed convolution we found this approach to deliver similar performance while preventing checkerboard artifacts in the network output. We then recombine the upsampled features with the features from the corresponding level of the context aggregation pathway via concatenation. Following the concatenation, a localization module recombines these features together. It also further reduces the number of feature maps which is critical for reducing memory consumption. A localization module consists of a 3x3x3 convolution followed by a 1x1x1 convolution that halves the number of feature maps.

Inspired by [9] we employ deep supervision in the localization pathway by integrating segmentation layers at different levels of the network and combining them via elementwise summation to form the final network output. Throughout the network we use leaky ReLU nonlinearities with a negative slope of $10^{-2}$ for all feature map computing convolutions. We furthermore replace the traditional batch with instance normalization [14] since we found that the stochasticity induced by our small batch sizes may destabilize batch normalization.

**Training Procedure** Our network architecture is trained with randomly sampled patches of size 128x128x128 voxels and batch size 2. We refer to an epoch as an iteration over 100 batches and train for a total of 300 epochs. Training is done using the adam optimizer [15] with an initial learning rate $\text{lr}_{\text{init}} = 5 \cdot 10^{-4}$, the following learning rate schedule: $\text{lr}_{\text{init}} \cdot 0.985^{\text{epoch}}$ and a l2 weight decay of $10^{-5}$.

One challenge in medical image segmentation is the class imbalance in the data that hampers the training when using the conventional categorical crossentropy loss. In the BraTS 2017 training data for example, there is 166 times as much background (label 0) as there is enhancing tumor (label 4). We approach this issue by formulating a multiclass Dice loss function, similar to the one employed in [11], that is differentiable and can be easily integrated into deep learning frameworks:

$$\mathcal{L}_{\text{dc}} = -\frac{2}{|K|} \sum_{k \in K} \frac{\sum_i u_{i,k} v_{i,k}}{\sum_i u_{i,k} + \sum_i v_{i,k}} \quad (1)$$

where $u$ is the softmax output of the network and $v$ is a one hot encoding of the ground truth segmentation map. Both $u$ and $v$ have shape $I$ by $K$ with $i \in I$ being the voxels in the training patch and $k \in K$ being the classes. $u_{i,k}$ and $v_{i,k}$ denote the softmax output and ground truth for class $k$ at voxel $i$, respectively.

When training large neural networks from limited training data, special care has to be taken to prevent overfitting. We address this problem by utilizing a large variety of data augmentation techniques. Whenever possible, we initialize these techniques using aggressive parameters that we subsequently attenuate over the course of the training. The following augmentation techniques were applied on the fly during training: random rotations, random scaling, random elastic deformations, gamma correction augmentation and mirroring.

The fully convolutional nature of our network allows to process arbitrarily sized inputs. At test time we therefore segment an entire patient at once, alleviating problems that may arise when computing the segmentation in tiles with a network that has padded convolutions. We furthermore use test time data augmentation by mirroring the images and averaging the softmax outputs over several dropout samples.

### 2.2 Survival Prediction

The task of survival prediction underpins the clinical relevance of the BraTS challenge, but at the same time is very challenging, particularly due to the absence of treatment information and the small size of the available dataset. For this subchallenge, only the image information and the age of the patients was provided.

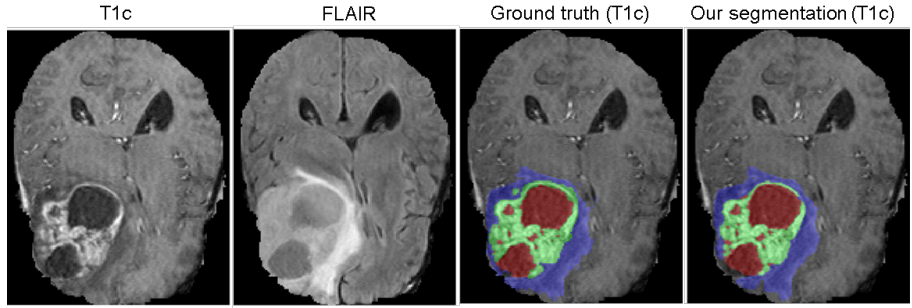

**Fig. 2.** Qualitative segmentation result. Our approach is capable of segmenting the large necrotic cores while also detecting the small structures within the tumor core. Edema is shown in blue, enhancing tumor in green and non-enhancing and necrotic tumor in red.

Our approach to survival prediction is based on radiomics. We characterize the tumors using image based features that are computed on the segmentation masks. We compute shape features (13 features), first order statistics (19 features) and gray level co-occurence matrix features (28 features) with the pyradiomics package [16]. The tumor regions for which we computed the features were the edema (ede), enhancing tumor (enh), necrosis (nec), tumor core (core) and whole tumor (whole). We computed only shape features for edema and the whole tumor, shape and first order features for tumor core and the entire feature set for non-enhancing and necrosis and enhancing tumor. With the image features being computed for all modalities, we extracted a total of 517 features.

These features are then used for training a regression ensemble for survival prediction. Random forests are well established in the radiomics community for performing well, especially when many features but only few training data are available. These properties make random forest regressors the prime choice for the scenario at hand (518 features, 163 training cases). We train a random forest regressor (RFR) with 1000 trees and the mean squared error as split criterion. Additionally, we designed an ensemble of small multilayer perceptrons (MLP) to complement the output of the regression forest. The ensemble consists of 15 MLPs, each with 3 hidden layers, 64 units per layer and trained with a mean squared error loss function. We use batch normalization, dropout ($p_{\mathrm{drop}} = 0.5$) and add gaussian noise ($\mu = 0, \sigma = 0.1$) in each hidden layer. The outputs of the RFR and the MLP ensemble are averaged to obtain our final prediction.

## 3  Results

### 3.1  Segmentation

We trained and evaluated our architecture on the BraTS 2017 and 2015 training datasets via five fold cross-validation. No external data was used and the networks were trained from scratch. Furthermore, we used the five networks obtained by the corresponding cross-validation as an ensemble to predict the respective validation (BraTS 2017) and test (BraTS 2015 and 2017) set. Both the training set and validation/test set results were evaluated using the online evaluation platforms to ensure comparability with other participants.

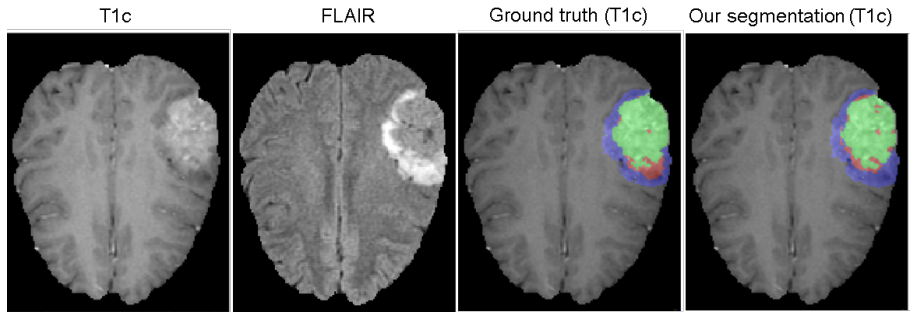

**Fig. 3.** The most prominent mode of error for the tumor core is the non enhancing tumor region. Edema is shown in blue, enhancing tumor in green and non-enhancing and necrotic tumor in red.

|  | Dice | | | Sensitivity | | | PPV | | |
|---|---|---|---|---|---|---|---|---|---|
|  | whole | core | enh. | whole | core | enh. | whole | core | enh. |
| Kamnitsas et al. [7] | **0.85** | 0.67 | 0.63 | 0.88 | 0.60 | 0.67 | **0.85** | **0.86** | 0.63 |
| Kayalibay et al. [9] | **0.85** | 0.72 | 0.61 | **0.91** | **0.73** | 0.67 | 0.82 | 0.77 | 0.61 |
| ours | **0.85** | **0.74** | **0.64** | **0.91** | **0.73** | **0.72** | 0.83 | 0.80 | **0.63** |

**Table 1.** BraTS 2015 test set results. We used the five models obtained by training a five fold cross-validation on the BraTS 2015 training data as an ensemble.

Table 1 compares the performance of our algorithm to other state of the art methods on the BraTS 2015 test set. Our method compares favorably to other state of the art neural networks and is currently ranked first in the BraTS 2015 test set online leaderboard. In Table 2 we show an overview over the segmentation performance of our model on the BraTS 2017 dataset.

Qualitative segmentation results are presented in Figures 2 and 3. Our network is capable of accurately segmenting large tumor regions (such as the necrotic cores in Figure 2) as well as fine grained details (scattered necrotic regions in the tumor core). Note how the thin wall of the enhancing region in the uppermost part of the tumor in Figure 2 was segmented with voxel-level accuracy whereas the manual ground truth label spilled into the bordering edema region. Furthermore, the small spot of enhancing tumor that is surrounded by edema in the ground truth segmentation can, upon closer inspection of the raw data (see patient Brats17_TCIA_469_1), be identified as a blood vessel that has been erroneously included in the enhancing tumor region by the annotator. Figure 3 demonstrates the main mode of error of our model. The former non-enhancing tumor label, which was integrated into the necrosis label for the BraTS 2017 challenge, is often not well defined in the training data. As a result, our algorithm learns where to predict this label from the context rather than based on image evidence and seems to sometimes guess where to place it.

| Dataset | Dice | | | Sensitivity | | | Specificity | | | Hausdorff Dist. | | |
|---|---|---|---|---|---|---|---|---|---|---|---|---|
| | whole | core | enh. | whole | core | enh. | whole | core | enh. | whole | core | enh. |
| BraTS 2017 Train | 0.895 | 0.828 | 0.707 | 0.890 | 0.831 | 0.800 | 0.995 | 0.997 | 0.998 | 6.04 | 6.95 | 6.24 |
| BraTS 2017 Val | 0.896 | 0.797 | 0.732 | 0.896 | 0.781 | 0.790 | 0.996 | 0.999 | 0.998 | 6.97 | 9.48 | 4.55 |

**Table 2.** Results for the BraTS 2017 dataset. Train: 5 fold cross-validation on the training data (285 cases). Val: Result on the validation dataset using the five models from the training cross-validation as an ensemble (46 cases).

Quantitatively, we achieve Dice scores of 0.896, 0.797 and 0.732 for whole, core and enhancing, respectively, on the BraTS 2017 validation set. This result places us among the best performing methods according to the online validation leaderboard. When comparing these values to the Dice scores achieved on the training set (0.895, 0.828, 0.707) we conclude that our model, together with the extensive data augmentation used during training, does not overfit to the training dataset. We purposefully did not submit more than once to the validation set in order to ensure that we do not overfit by adapting our hyper parameters to the validation data.

| | Dice | | |
|---|---|---|---|
| | enh. | whole | core |
| Mean | 0.647 | 0.858 | 0.775 |
| StdDev | 0.326 | 0.161 | 0.269 |
| Median | 0.795 | 0.910 | 0.886 |
| 25 quantile | 0.619 | 0.856 | 0.764 |
| 75 quantile | 0.863 | 0.940 | 0.932 |

**Table 3.** BraTS 2017 test set results. The scores were computed by the organizers of the challenge based on our submitted segmentations.

Table 3 shows the test set results as reported back to us by the organizers of the challenge. We achieved mean Dice scores of 0.858 (whole tumor), 0.775 (tumor core) and 0.647 (enhancing tumor). These scores are lower than the ones obtained on either training or validation set, which is surprising provided that we did not observe overfitting during training and on the validation set. Based on the high median Dice scores we hypothesize that the test set contained a significant number of very difficult cases. Also, we are uncertain how cases with no enhancing tumor in the ground truth segmentation are aggregated into the mean since their Dice score is always zero by definition.

| Features | Ground Truth Segmentation | | | Our Segmentation | | |
|---|---|---|---|---|---|---|
| | RFR | MLP ens | combined | RFR | MLP ens | combined |
| shape, age (66) | 344.89 | **352.00** | **339.61** | 353.12 | **343.19** | **335.08** |
| glcm, age (225) | 348.14 | 462.16 | 381.25 | 350.78 | 388.99 | 357.41 |
| first order, age (229) | 358.69 | 388.44 | 362.20 | 354.66 | 381.42 | 355.89 |
| shape, glcm, age (290) | **344.86** | 431.96 | 367.14 | **346.40** | 378.73 | 349.13 |
| shape, first order, age (294) | 352.64 | 372.59 | 350.62 | 351.56 | 360.24 | 342.46 |
| glcm, first order, age (453) | 353.18 | 443.64 | 378.83 | 354.30 | 383.82 | 356.25 |
| all (518) | 350.40 | 385.66 | 354.86 | 352.95 | 372.04 | 348.55 |

**Table 4.** Survival prediction experiments. We trained a random forest regressor (RFR) and a MLP ensemble (MLP ens). Averaging the regression outputs of the RFR and MLP ensemble yields the *combined* result. The best root mean squared error is achieved when using RFR and MLP ensemble together with only shape features and the patients age.

### 3.2 Survival Prediction

We extensively evaluated the components of our regression ensemble as well as different feature sets with the aim of minimizing the mean squared error by running 5-fold cross-validations on the 163 provided training cases. A summary of our findings for both the ground truth and our segmentations is shown in Table 4. We observed that the random forest regressor performs very well across all feature sets while the MLP ensemble is much less stable with an increasing number of features. The overall best results were obtained by averaging the MLP ensemble output with the one from the random forest regressor (column *combined*) and using only shape features and the age of a patient. Interestingly, while the random forest performance is almost identical between ground truth and our segmentations, the MLP ensemble performs better on our segmentations for all feature sets, which is also reflected by the *combined* results. The best root mean squared error we achieved was 335.08 (mean absolute error 232.76) in a five-fold cross-validation on the training set. On the test set we obtained 457.83 RMSE (MSE 209607), an accuracy of 52.6% and a Spearman correlation coefficient of 0.496.

## 4 Discussion

In this paper we presented contribution to the BraTS 2017 challenge. For the segmentation part of the challenge we developed a U-Net inspired deep convolutional neural network architecture which was trained from scratch using only the provided training data, extensive data augmentation and a dice loss formulation. We achieve state of the art results on BraTS 2015 and presented promising scores on the BraTS 2017 validation set. On the test set we obtained mean dice scores of 0.858 for whole tumor, 0.775 for tumor core and 0.647 for the contrast enhancing tumor. Training time was about five days per network. Due to time restrictions we were limited in the number of architectural variants and data augmentation methods we could explore. Careful architecture optimizations already allowed us to train with large 128x128x128 patches and a batch size of 2 with 16 filters at full resolution, which is significantly more than in [9]. Training with larger batch sizes and more convolutional filters in a multi-GPU setup should yield further improvements, especially provided that we did not observe significant overfitting in our experiments. While most of our effort was concentrated on the segmentation part of the challenge, we also proposed an ensemble of a random forest regressor and a multilayer perceptron ensemble for the survival prediction subchallenge. By using only shape based features, we achieved a root mean squared error of 335.08 and a mean absolute error of 232.76 in a five fold cross-validation on the training data and using our segmentations. On the test set, our survival prediction approach obtained 457.83 rmse (209607 mse), an accuracy of 52.6% and a Spearman correlation coefficient of 0.496. The survival prediction task could be improved further by considering the position in the tumor relative to other structures in the brain such as the ventricles, optical nerve fibers or other important fibre tracts. Furthermore, our group based manual feature selection should be replaced by a proper feature selection algorithm such as forward/backward selection [17] or a feature filter based approach [18].